# Privileged Information for Data Clustering

Jan Feyereisl[a,*], Uwe Aickelin[a]

[a]*School of Computer Science, The University of Nottingham, UK*

**Abstract**

Many machine learning algorithms assume that all input samples are independently and identically distributed from some common distribution on either the input space $X$, in the case of unsupervised learning, or the input and output space $X$ x $Y$ in the case of supervised and semi-supervised learning. In the last number of years the relaxation of this assumption has been explored and the importance of incorporation of additional information within machine learning algorithms became more apparent. Traditionally such fusion of information was the domain of semi-supervised learning. More recently the inclusion of knowledge from separate hypothetical spaces has been proposed by Vapnik as part of the supervised setting. In this work we are interested in exploring Vapnik's idea of 'master-class' learning and the associated learning using 'privileged' information, however within the *unsupervised* setting.

Adoption of the advanced supervised learning paradigm for the unsupervised setting instigates investigation into the difference between privileged and technical data. By means of our proposed aRi-MAX method stability of the K-Means algorithm is improved and identification of the best clustering solution is achieved on an artificial dataset. Subsequently an information theoretic dot product based algorithm called *P-Dot* is proposed. This method has the ability to utilize a wide variety of clustering techniques, individually or in combination, while fusing privileged and technical data for improved clustering. Application of the *P-Dot* method to the task of digit recognition confirms our findings in a real-world scenario.

*Keywords:* Clustering, Privileged Information, Hidden Information, Master-Class Learning, Machine Learning

## 1. Introduction

At the core of machine learning lies the analysis of data. Data are worthless unless they contain meaningful information and thus useful knowledge about a particular problem or a set of problems. For different areas of machine learning we can categorise data based on what we know about them, before they are subject to a particular algorithm. The three core types of learning, supervised, semi-supervised and unsupervised learning, differ first and foremost in the type of data they have at their disposal. We highlight these differences in Table 1.

In the *supervised* setting a set of $n$ examples $X = (x_1, ..., x_n)$ is provided, along with a set of labels $Y = (y_1, ...,y_n)$, resulting in a set of pairs of observations $S = (x_1, y_1), ..., (x_i, y_i)$. In the *semi-supervised* setting, the same type of information is available, however commonly with only a small subset of examples $X_l \subset X$ with corresponding labels $Y_l \subset Y$. In contrast to the supervised setting, the amount of unlabelled examples $X_u \subset X$ can be fairly large. In addition, or sometimes instead of the subset $X_l$ of labelled examples, a set of constraints can exist that can be imposed upon the employed algorithm. Such constraints traditionally denote whether a pair or a set of points should or should not co-exist in the same cluster. In *unsupervised* learning the amount of knowledge about data to be analysed is the most restrictive. Only a set of $n$ examples from $X$ is supplied. This makes unsupervised learning hard to define formally and thus a difficult computational problem [41].

One aspect that all three types of learning share is the fact that all sample points should be selected independently and identically distributed from some common distribution on either $X$, in the case of unsupervised learning, or $X$ x $Y$ in the case of supervised and semi-supervised learning. The restriction on the distribution from which the input samples are collected has increasingly been relaxed within the literature and its consequences explored [6]. Importance of incorporation of additional information within machine learning algorithms became more apparent with the introduction of multiple view learning [1, 8] and learning using privileged information (LUPI) [37]. Traditionally such fusion of knowledge was the domain of semi-supervised learning, where techniques such as co-training [3] were employed in order to fuse separate data in order to exploit knowledge encoded within unlabelled data. More recently the inclusion of knowledge from separate hypothetical spaces has also been proposed by Vapnik [36, 38, 37] as part of the supervised setting. In his work the notion of *"privileged"* and *"hidden"* information denotes the existence of an additional set of data that provides a higher level information, akin to information provided by a *"master"* to a pupil, about a specific problem. In the supervised setting such information is only available during training. The fusion of separate hypothetical spaces for the purpose of unsupervised learning, particularly cluster analysis has been investigated in the past [2, 13, 10], however not within the LUPI framework.

---

[*]Corresponding Author
*Email address:* jqf@cs.nott.ac.uk (Jan Feyereisl)

Table 1: Differences in input knowledge across the three types of learning.

| Learning | Data | | |
|---|---|---|---|
| | X | Y | Other |
| Supervised | $X = (x_1, \ldots, x_n)$ | $Y = (y_1, \ldots, y_n)$ | – |
| Semi-Supervised | $X_l = (x_1, \ldots, x_l)$ | $Y_l = (y_1, \ldots, y_l)$ | – |
| | $X_u = (x_{l+1}, \ldots, x_{l+u})$ | – | Constraints |
| Unsupervised | $X = (x_1, \ldots, x_n)$ | – | – |

In this work we are interested in exploring the idea proposed by Vapnik, however within the *unsupervised* setting. We are particularly interested in the notion of 'master-class' learning and the associated learning using 'privileged' information which is explained in Section 2. Section 3 provides insights into the difference between information as meant in the traditional sense and the so-called 'privileged' information. In Section 4 the question of whether such information can be used to improve data clustering is investigated and a method for combining 'privileged' information as part of a clustering solution is proposed. Section 5 highlights the use of our method on a real world dataset. The paper concludes with Section 6 where our results are summarised and future work is proposed.

## 2. Learning Using Privileged Information

To understand the notion of learning using privileged information, first the wider context of 'learning from empirical data' is depicted. In machine learning, supervised learning is a subset of learning techniques that have one common goal. This goal is to learn a mapping from input *x* to an output *y*. The standard input for supervised techniques consists of a set $X = (x_1, \ldots, x_n)$ of *n* examples from some space $\chi$ of interest. Typically this set of examples is drawn independently and identically distributed (i.i.d.) from some fixed but unknown distribution with the help of a generator (Gen). Along with such examples we are also given a set $Y = (y_1, \ldots, y_n)$ of labels $y_i$ that correspond to our examples $x_i$. This set is said to have been created by a supervisor (Sup), who knows the true mapping from *x* to *y*. Thus we are provided with a set of pairs $(x_1, y_1), \ldots, (x_i, y_i)$ and from these we aim to learn the real mapping as accurately as possible using our learning machine (LM). The general model of learning from examples, adopted from Vapnik [39], can be seen in Figure 1(a). In this figure a generator samples data *x* i.i.d. from the unknown distribution of a given problem, which is subsequently paired with an appropriate label *y* by the supervisor. The pair (*x*, *y*) is then used by the learning machine to learn the mapping from *x* to *y* in order for the machine to be able to give as similar an answer, to the supervisor, as possible.

Figure 1(b) displays the concept of learning using *'privileged information'*, pertinent to our investigation. In comparison to the supervised setting, there exists an additional data generator $Gen*_{priv}$ of input data $x^*$. This generator is different from the only generator that exists in the traditional supervised setting and which in this figure is called $Gen_{tech}$. Existence of two separate generators suggests that the inputs *x* and $x^*$ do not need to come from the same distribution. It is however important that they come from the same domain, i.e. the domain of the problem that we attempt to solve or learn about.

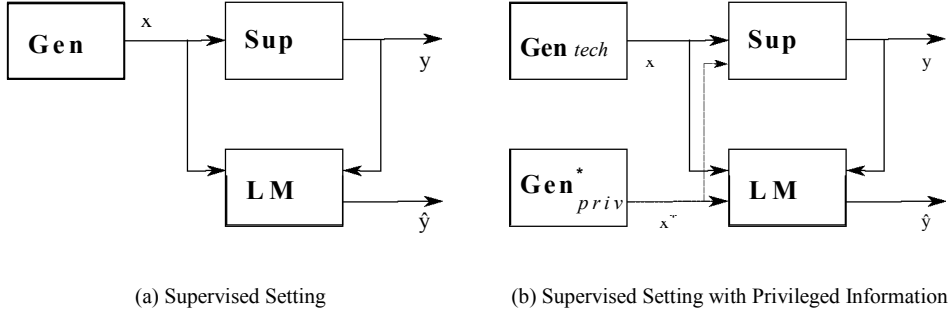

(a) Supervised Setting      (b) Supervised Setting with Privileged Information

Figure 1: Two models of learning from examples. In the LUPI setting (b) an additional generator of data $x^*$ exists. This data is called privileged as it is available only during training.

Originally, Vapnik [36] suggested a learning paradigm called *'master-class'* learning, where a teacher plays an important role. This teacher is not the supervisor (Sup) as used in a supervised setting. The teacher is the additional data generator $Gen^*priv$. In this paradigm the teacher is an entity that provides information akin to information provided by a human teacher. The teacher provides students with *hidden information*, $(x^*)$. This information is not apparent at first or explicitly stated. It is usually hidden within the actions of the teacher [38]. It provides the students with the teacher's view of the world. It is available to the students in addition to the information that exists in textbooks and as a result they can learn better and faster. It is important to note that the notion of 'hidden information' is di*ff*erent from information hiding in the the field of data hiding [9, 34]. Both areas deal with information passed across a (possibly) covert channel, the purpose and use of such information is however di*ff*erent in both cases.

In [37], Vapnik renamed his learning paradigm to *'Learning Using Privileged Information'*. In this formulation the notion of privileged rather than hidden information is presented, where the data is said to be privileged as it is available to us only during training and not during testing.

To realize this advanced type of learning, Vapnik developed the SVM+ algorithm [38], where the fusion of privileged information with the classical, technical, data is performed. To understand how this fusion works we refer to the SVM decision function, shown below:

$$f(x) = (w \cdot z) + b = \sum_{i=1}^{n} \alpha_i y_i K(x_i, x) + b \quad (1)$$

In the SVM+ method this decision function depends on the kernel $K$ defined in the transformed feature space, however coe*ffi*cients $\alpha$ depend on both the transformed feature space as well as on a newly defined correction space $\varphi(x^*)$. The correction space is where the privileged data is optimised and thus incorporated as part of the overall solution. Thus in addition to the above decision function, an additional correcting function was introduced [38]:

$$\varphi(x^*_j) = (w^* \cdot z^*_i) + d = \frac{1}{\gamma} \sum_{i=1}^{Z} (\alpha_i + \beta_i - C) K^*_{i,j} + d \quad (2)$$

An important strength of the SVM+ algorithm is the ability of the system to reject privileged information in situations when similarity measures in the correcting space are not appropriate, thus privileged information is only used when it is deemed beneficial. One drawback of the system is the increase in computational requirements due to the necessity of tuning of more parameters than in the original SVM setting [37].

Experimental results using the above algorithm show the new paradigms' superiority in terms of performance over the original SVM method. Vapnik shows that a poetic description [38] of a set of images of numbers provides more useful knowledge for learning than knowledge embedded within a higher resolution image, which holds more "technical" information about the underlying digits. In Vapnik's work a poetic description is a poet's textual depiction of the underlying image, described in section 5.1. In [37] the work is extended to show its success in tackling a bio-informatics and a time-series prediction problem.

More recently Pechyony has analysed the LUPI paradigm theoretically [25, 26]. The LUPI paradigm has also been compared to the problem of structured or multi-task learning in both the classification [20] as well as the regression settings [5]. The multi-task learning framework considers problems where training data can naturally be separated into several groups, which can in turn be used to perform a number of individual model selections. In [20], the authors suggest that the LUPI setting is a similar problem, where training data are structured, however used to create only a single model.

*2.1. What is Privileged Information?*

To understand the problem that is to be solved in this work, first a description of Vapnik's *"privileged information"* needs to be given. Here we will compare it to data as considered in the traditional sense. Vapnik named this traditional data *"technical data"*, as in most cases such data originated from a technical process, such as a pixel space in the case of a digit recognition task or amino-acid space in the case of classification of proteins. To help us understand what "privileged information" is, it is useful to present examples where such information can become useful. Vapnik suggested three example types of privileged information [37]:

*Advanced Technical Model:*. In this scenario the privileged information can be seen as a high level technical model of a particular problem to be solved. An example of such a model is the 3D-structure information of proteins and their position within a protein hierarchy in the field of bioinformatics. This 3D-structure is a technical model developed by scientists to categorise and classify known proteins. Technical data on the other hand refers to amino-acid sequences on which classification is performed using most traditional approaches. When information contained within the known 3D-structures can be used to improve learning performed on the amino-acid sequences, without the 3D-structures being required for future predictions, privileged information becomes useful.

*Future Events:*. Many computational problems involve the prediction of a future event, given a set of current measurements. An example of privileged information in this scenario is a set of information provided by an expert in addition to the set of current measurements. For instance if the task at hand is the prediction of a particular treatment of a patient in a year's time, given his/her current symptoms, a doctor can provide information about the development of symptoms in three, six and nine months time.

*Holistic Description:*. The last example type of privileged information relates to holistic descriptions [29] of specific problems or problem instances by entities that are associated to the problem domain. Considering a medical problem again where, in this case, biopsy images are to be classified between cancerous and non-cancerous samples, technical data are the individual pixels of each image. Privileged information on the other hand are reports written about the images by pathologists in a high-level holistic language. The aim of the computational task becomes the creation of a classification rule in the pixel space with the help of the holistic reports produced by pathologists, so as to allow for future classifications of biopsy images without the need of a pathologist.

The above three example types of privileged information are only a very small selection of the possible set of additional information that could be obtained from a number of problem domains. Vapnik states that almost any machine learning problem contains some form of privileged information, which is currently not exploited in the learning process [37]. This includes the unsupervised learning process that we are interested in tackling.

## 3. The Difference Between Information and Privileged Information

The notion of 'privileged' information accentuates the question of what 'privileged' information actually is and why it should be treated differently than other types of data. In the previous paragraph we have highlighted a number of examples of privileged information. In this section we will first show that there is a difference in the type of data that one can obtain for a particular problem to be solved. More specifically we will show that there is a difference between traditional feature space data and the so-called 'privileged' information. We will demonstrate this by observing differences in the results of the K-Means clustering algorithm on various mixtures of these two types of data. Subsequently we will postulate that the method of fusion of these differing input data has an impact on the data's contribution towards a better solution in the unsupervised learning setting and cluster analysis in particular.

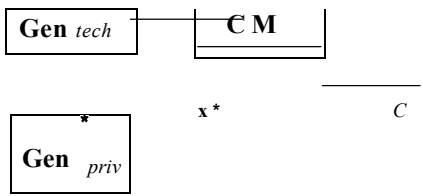

Figure 2: Unsupervised Learning Machine using Privileged Information. In this setting the clustering machine (CM) allows for the fusion of information from both data sources for the purpose of improved clustering.

*3.1. Privileged Information in Supervised Learning*

In the supervised setting, Vapnik [36, 38, 37] has shown that a learning machine trained with the help of both privileged information, as well as the original technical data, provides improved performance over a machine trained only on technical data. He has also shown that if privileged information is available, the technical data does not need to be of as high quality as when only technical data is used for training. To put this into context, Vapnik compared the classification performance of a learning machine (SVM+) trained on low resolution images of digits with privileged information against a learning machine trained on high-resolution images of the same digits. This comparison has shown comparable results. It is however not known whether privileged information would provide similar type of performance increase when used directly as additional features. Below we demonstrate that there is a difference in combining data from different spaces in ways other than simply concatenating privileged information as additional features to the technical data.

*3.2. Privileged Information in Unsupervised Learning*

The use of any type of privileged information as part of the unsupervised setting has not yet been performed. In order to show the importance of privileged information, we will highlight its usefulness in the domain of cluster analysis. Figure 2 shows, with the help of the learning machine framework, the unsupervised learning setting using privileged information.

Similarly to Figure 1(b), we have two generators, where $Gen_{tech}$ is the generator of the original technical data and $Gen^*_{priv}$ supplies the learning machine, in this case the Clustering Machine *CM*, with privileged information. Unlike in the supervised setting there is no supervisor, *Sup*, thus the clustering machine needs to be able to exploit the information encoded within the two data sources, without the knowledge of the number of classes or which instances belong to which group. The clustering machine produces a clustering *C*, which provides a set of meaningful partitions of *X* according to information encoded in $x$ and $x^*$.

*3.3. Clustering Using Perfect Privileged Information*

To demonstrate the usefulness of privileged information, we designed an experiment that highlights the difference between privileged information as a separate source of data and privileged information as a set of additional features. We created an artificial dataset that consists of a clear example of a problem with which any clustering algorithm has difficulties dealing with. The dataset can be seen in Figure 3. This figure shows the original technical data. This dataset is symmetric in the distribution of points, however it is asymmetric in terms of class assignment. This problem is ill-posed as it likely violates the *cluster assumption* and *low-density separation assumption* [7]. For this reason no clustering algorithm is able to solve it. Four possible solutions for the problem, using the K-Means algorithm, can be seen in Figure 4. These solutions are all incorrect and represent approximately 80% of all solutions produced by K-Means, depending on the starting locations of the centres of the algorithm. The only possibility for improving the quality of the clustering solution is to obtain additional information. In the real world, such information may be difficult to obtain, however in many situations where information from the same source cannot be obtained, information from the same domain can be obtained instead with the help of an expert of some kind.

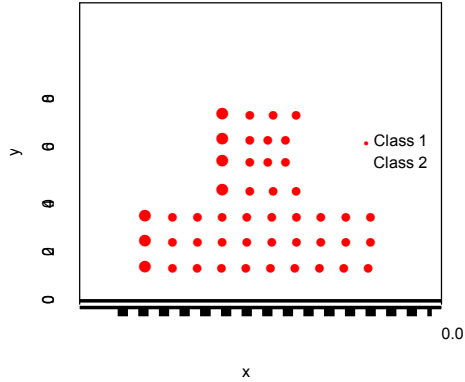

Figure 3: Artificial dataset with true class assignment shown. The dataset is symmetric in point distribution however asymmetric in class assignment. This problem cannot be solved using standard clustering techniques.

In our case we first assume a *perfect* expert that can supply information akin to class labels in terms of class separability. By this we mean that the additional information allows for clear separation of the dataset into the correct true clusters. However as this is an unsupervised setting, this additional information cannot be used in the same sense as labels in a supervised setting. Unlike labels, this information cannot be used with a correcting function as there are no guarantees about the correctness of the privileged information and no information about which cluster belongs to a particular class. The information chosen by us for this purpose can be seen in Figures 5(a) and 5(b). Two sets of privileged information were chosen. This information, which we termed *point-wise*, as all points belonging to a class are located at the same location, can be thought of as two dimensional set of points, where each point is associated with an existing data instance of a particular class in the technical dataset. The two sets of data only differ in the Euclidean distance, $d$, also denoted by the Euclidean norm $\|.\|$, between points that are representative of the two classes. For simplicity and clarity, privileged information for data items in dataset where the distance between the two classes is $d = \|x^*_1 - x^*_2\| = 0.2$, that belong to class *one* (●), are *all* at location (0.1,0.1), whereas for class *two* (▲), privileged information is a set of points *all* located at (0.2,0.2). In the second privileged dataset where $d = \|x^*_1 - x^*_2\| = 0.5$, points belonging to class *one* (●) are *all* at location (0.1, 0.1) and to class *two* (▲) at (0.5,0.4). These two different types of data were chosen to reflect on the fact that the larger the difference between values in different classes, the more separated the two groups become in that particular dimension. Thus if our additional information is very well separated due to $d$ being very large, then in some cases the problem of separating the two groups becomes easier when concatenating this data in the original feature space. However if $d$ is small, with respect to other values present in the technical dataset, then even if such attribute is vital for the successful clustering, its influence on the solution will be minimal, especially when the number of dimensions of the technical data set is large with respect to the number of dimensions of the additional data.

*3.3.1. Adjusted Rand Index*

A clustering validity measure is required to evaluate the performance of the clustering solution provided by all tested clustering algorithms. A method called the adjusted Rand index [17] was employed to assess the similarity between the clustering solutions produced by the clustering algorithms and the true solution. This method is a corrected-for-chance version of the RAND index [28], which assesses the degree of agreement between two partitions of the same set of data. The method has a maximum value of 1, which denotes a perfect agreement between the two tested partitions. The minimum of 0 denotes that the two sets do not agree on any pairs of points.

*3.3.2. Clustering of X Concatenated with X - (X + $X^*$)*

In the first experiment the K-Means algorithm is applied to the original dataset, comprising only of technical data, $X$. An agreement between the true clustering and the clustering performed by K-Means using the adjusted Rand index method is calculated across 100 runs of the algorithm. As stated earlier the results of the first experiment confirm that the algorithm cannot find the correct solution and that in many cases it finds a completely wrong solution. Concordance of 62% between the K-Means solution and the true class assignment is found in the best case, with a mean concordance of 52%. On the other hand the worst solution finds no agreement between the true clustering and the solution that K-Means found. This is due to the fact that the algorithm is susceptible to local optima, which are typically found when the initial centres of the algorithm are not chosen optimally. As the K-Means algorithm traditionally selects such initial centres at random, it is understandable that many of the algorithm's solutions might be incorrect. Numerous solutions to this initialisation problem of the K-Means algorithm have been proposed in the past, for example [4, 23, 21], however we are not interested in solving this problem per se, nevertheless our analysis does have implications that are related.

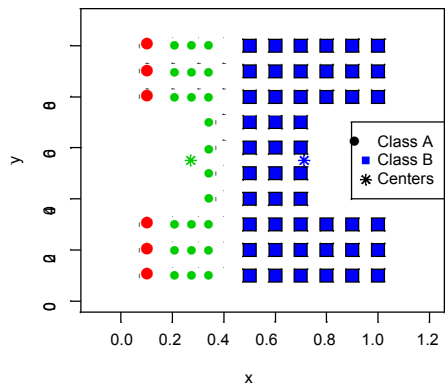
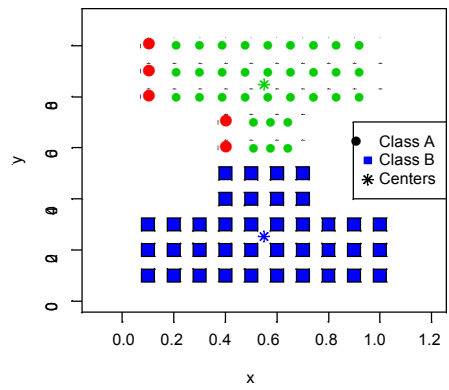
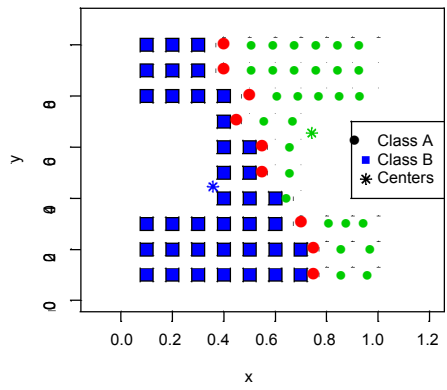
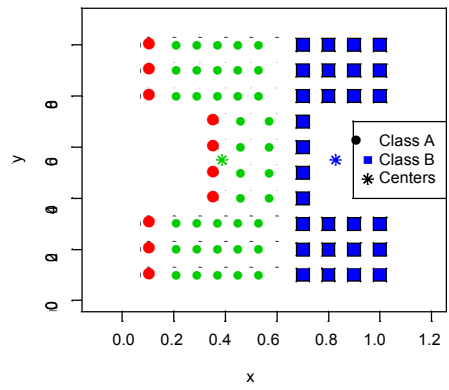

(a) Solution 1

(b) Solution 2

(c) Solution 3

(d) Solution 4

Figure 4: Application of the K-Means clustering algorithm on the artificial dataset and its failure due to local optima. The symmetric nature of the dataset provides ambiguity with respect to how the data should be partitioned. Thus many clustering algorithms, such as the K-Means algorithm, will fail in correctly partitioning the data into the appropriate clusters.

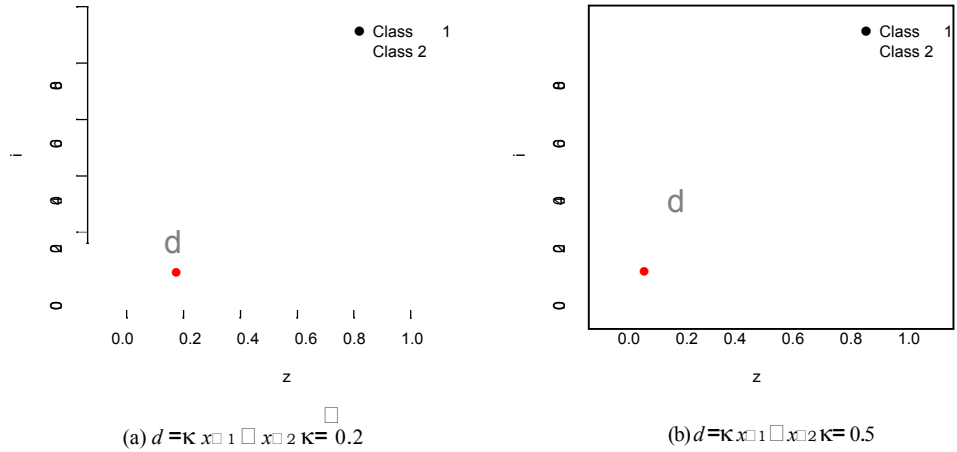

(a) $d = \kappa x_1 - x_2 \kappa = 0.2$  (b) $d = \kappa x_1 - x_2 \kappa = 0.5$

Figure 5: Point-wise privileged information, $x^*$, supplied in addition to original technical data, $x$. Here $d$ denotes the Euclidean distance between the points in the two classes $x_1$ and $x_2$. Each class contains as many items in $x^*$ as there are items in $x$, however as these are all located at the same position in this scenario, only one item per class can be seen in this figure.

Once the results obtained on the original technical data, $X$, show that the K-Means algorithm cannot achieve a very good solution, we now include the privileged information as a part or in addition to the original feature space.

In the first instance we obtain results for the clustering produced by K-Means when the privileged information is fused into the feature space of the technical data. Hence the additional data is simply *appended* to the original dataset, and thus can be thought of as only an additional set of attributes. We will refer to this scenario throughout the rest of this paper using the following notation: $X + X^*$, where $+$ denotes a concatenation of the two sets of data, resulting in one dataset. To evaluate whether significant differences exist in the results, we examine our results with the help of a statistical test called the Wilcoxon signed-rank test [42]. A confidence interval level of 0.05 was chosen and a null hypothesis with two alternative hypotheses was given.

By including the additional information as simply another set of attributes we observe that across 100 runs the quality of solutions of the K-Means algorithm in the case when $d = 0.2$ slightly decreases, from 0.52 to 0.51, when looking at the mean adjusted Rand index value. Comparing the clustering of $X$ with clustering of $X + X^*$ statistically, we cannot reject any of our proposed three hypotheses. The null hypothesis states that the two results are the same. The two alternative hypotheses state that the distribution of either one or the other result has a shift to the right with respect to the distribution of the other result. The statistical analysis suggests that the two results are not statistically significantly different. Thus there is no evidence for the benefit of using the additional information from $X^*$ by combining it with $X$ for the case when $d = 0.2$.

For the case where $d = 0.5$, the change in results is however substantial. The average agreement between the two clusterings improves dramatically, from 0.52 to 0.97. We also note that in this situation when the algorithm finds a good solution, it is a correct solution. Thus in this case we have an improvement in the clustering result when additional information helps to sufficiently separate the two groups, making correct solutions possible. In this case we cannot, however, call this additional information privileged because it is simply an additional set of features. Here the fusion of $X + X^*$ provides enough separation that influences the K-Means algorithm by a sufficient amount. The two attributes from the technical space now comprise only 50% of the information based on which the K-Means algorithm makes a decision. The other 50% comprises of the additional information which can be thought of as perfect with a high level of separability due to $d = 0.5$. This result highlights the importance of separability of classes based on the separability of individual features within an analysed dataset. As our original dataset comprises of two dimensions and the privileged dataset is also two dimensional, if the privileged dataset on its own is well separated, this has great influence on the analysis of the combined dataset, especially in cases where the dimensionality of $X$ does not differ greatly from the dimensionality of $X^*$. In cases where the dimensionality of the technical data is substantially larger than the dimensionality of $X^*$, then even if $X^*$ is perfectly linearly separable, the data in $X$ will degrade the influence of the information in $X^*$.

It is also important to note that if our data is always as clearly separable as in this example, then we can always obtain very good solutions using traditional methods. The above optimal case is however rarely to be experienced. Furthermore, distinct separation is difficult to obtain if our privileged information includes substantial noise, similar to that of the technical data. A question begs whether the fact that our privileged information comes from the same domain, rather than the same distribution, can be of benefit. In answering this question and help us overcome the above mentioned issue of dimension dependent separability, we used privileged information in a way that fuses knowledge from the two separate data sources, however not in the feature space of the technical data.

*3.3.3. Clustering Consensus of Disparate Hypothetical Spaces - aRi-MAX*

As both technical data and privileged data essentially come from the same domain, they should provide an insight into the problem that we are trying to analyse. In terms of clustering, the two datasets should provide information about the same or similar sets of clusters that belong to the given domain. Assuming the above is true, when each dataset is clustered individually, we should be able to achieve a consensus of the two results. Consequently we should understand better the underlying structure of the analysed data. The field of consensus clustering, sometimes also called aggregation of clustering [33, 35, 14], addresses a similar problem, where a number of clusterings of the same dataset exists and a consensus between those is sought after. Here we are interested in the clustering of two datasets that come from the same domain but different generators. In this section we use the clustering of the privileged information to select the best possible clustering of the technical data. We have devised an algorithm, called *aRi-MAX*, for this purpose. Pseudo-code for *aRi-MAX* is shown in Algorithm 1.

---
**Algorithm 1:** aRi-MAX - Clustering consensus of $X$ and $X^*$ using adjusted Rand index.

**Input**: Technical Data $X$, Privileged Data $X^*$
**Output**: Clustering of $X$, maximizing agreement between $X$ and $X^*$

1 initialization;
2 **foreach** *i in Runs* **do**
3   | $Ct_i \leftarrow$ clust($X$);
4   | $Ce_i \leftarrow$ clust($X^*$);
5 **end**
6 **foreach** *i in $Ct_i$* **do**
7   | $Cc_i \leftarrow$ max(adjustedRandIndex($Ct_i,Ce$))
8 **end**
9 max($Cc_i$)

---

In *aRi-MAX* the two datasets are fused by a consensus of clusterings that are evaluated using the adjusted Rand index method. First, a number of clustering solutions of the technical data are performed, followed by a number of clustering solutions of the privileged information. This corresponds to steps **2-5** in algorithm 1, where `clust(x)` can be any type of clustering algorithm. In our experiments this is the K-Means algorithm. Subsequently solutions of these two clusterings are compared using the adjusted Rand index method and the two cluster solutions with the

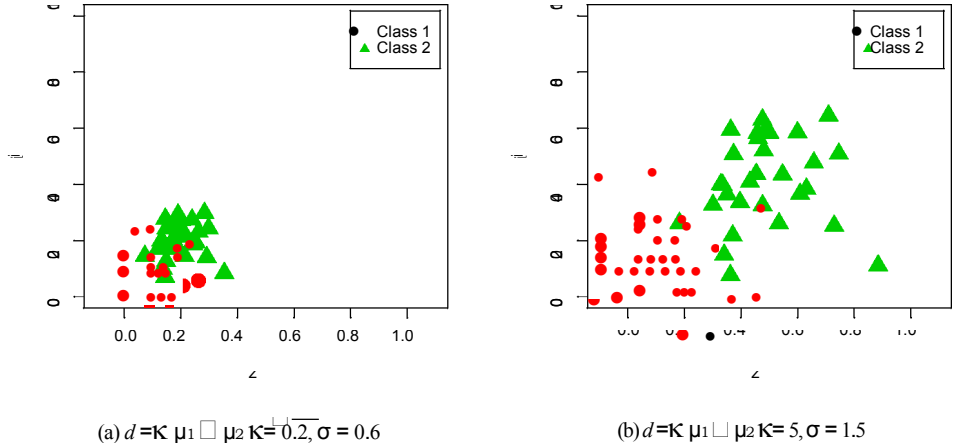

(a) $d = \|\mu_1 - \mu_2\| = 0.2, \sigma = 0.6$  (b) $d = \|\mu_1 - \mu_2\| = 5, \sigma = 1.5$

Figure 6: More realistic, Gaussian noise based privileged information, $X^*$. This privileged information is more akin to data in real-world scenarios, where numerous data items from different classes overlap, making a separation of the classes more difficult. $d$ again denotes the Euclidean distance between the centres of the two existing classes and $\sigma$ denotes the standard deviation.

highest agreement are selected as the final solution candidate, steps **6-9** in 1. Once this is completed, the candidate clustering solution of the technical dataset is evaluated against the true solution.

### 3.3.4. Clustering by Fusion of X with $X^*$ - $(X X^*)$

The fusion of $X$ and $X^*$ by means of our *aRI-MAX* method as well as our future fusion methods shall be denoted in the following way, $X X^*$. Here however the term shall not be interpreted in the strict mathematical sense denoting congruence. The choice of this symbol was due to the fact that $X$ and $X^*$ essentially should provide a congruent view of the same problem domain.

Experiments using the *aRI-MAX* method were performed on $X X^*$, for both the $d = 0.2$ and the $d = 0.5$ cases. Results obtained show that across the one hundred runs we are now able to obtain the best achievable result, given the dataset and clustering algorithm, 100% of the time in both cases. Thus the average agreement between the correct clustering and our solution across the 100 runs is 0.62. Even though the performance of our *aRi-MAX* method is limited by the capability of the underlying clustering technique applied on the technical data, $X$, in the case of K-Means, the variability of the outcome of the algorithm is reduced from a standard deviation ($\sigma$) of 0.23 for the $d = 0.2$ case and 0.1 for the $d = 0.5$ case to $\sigma = 0$ in both cases when our *aRI-MAX* method is used. Statistical tests performed reveal that obtained results are statistically significant and that the distribution of the results of our method were shifted to the right from the distribution of the results obtained on both the original data, $X$, as well as on the combined data, $X + X^*$, fused in feature space in the $d = 0.2$ case. These results confirm that the use of privileged information, in a way different than as part of the original feature space, is a viable direction.

### 3.4. Clustering Using Imperfect Privileged Information

The previous example was a very simplified form of the problem we are trying to tackle. In the next step, we investigate a situation, in which our privileged information is slightly more realistic. Additional information is modelled using Gaussian noise, centred on the location of privileged information used in the previous experiment. The new dataset is depicted in Figure 6. To investigate the amount of information that the privileged information holds about the underlying problem, the privileged data was clustered using K-Means on its own. Results show that the privileged data by itself can at best reveal the underlying clusters with approximately 62% to 66% accuracy in the $d = 0.2$ case and with 80% accuracy in the $d = 0.5$ case. Thus on its own, the data in $X^*$ is not any more useful than the technical data, $X$, for $d = 0.2$. For $d = 0.5$, the additional 20% might provide a benefit when the privileged information is fused by $X + X^*$, similarly to the point-wise dataset with $d = 0.5$.

Table 2: Results of statistical analysis of performance of K-Means algorithm on technical data, *X*, versus fusion $X + X^*$ and the *aRi-MAX* method on $X = X^*$. *Gaussian* privileged information with $d = ||\mu_1 - \mu_2|| = 0.2$. Statistical test performed is Wilcoxon signed-rank test at the 0.05 confidence interval level.

| | Clustering | | | $R_1 = R_2$ | | $R_1 < R_2$ | | $R_1 > R_2$ | |
|---|---|---|---|---|---|---|---|---|---|
| $R_1$ | | | $R_2$ | p-value | reject? | p-value | reject? | p-value | reject? |
| | $X$ | vs. | $X + X^*$ | 0.08318 | no | 0.04159 | yes | 0.9599 | no |
| | $X$ | vs. | $X = X^*$ | 0.00024 | yes | 0.9999 | no | 0.00012 | yes |
| | $X + X^*$ | vs. | $X = X^*$ | 4.207e-05 | yes | 1 | no | 2.104e-05 | yes |

*3.4.1. Clustering of X Concatenated with $X^*$ - ($X + X^*$)*

When this more noisy type of additional knowledge is fused with the technical data in the original feature space, $X + X^*$, in the $d = 0.2$ case, the results of the K-Means clustering algorithm slightly degrade to 48% compared to using only technical data, *X*, where a result of 52% is achieved. Again in the case where $d = 0.5$, the overall performance improves, however this time not as dramatically as in the point-wise dataset case with an average concordance of 91% rather than 97%. Statistical tests were also performed on the results obtained from the new fusions of data using Gaussian privileged information. From these tests it is apparent that for $d = 0.2$ we reject the hypothesis that the distribution of the result of the fused $X + X^*$ space is shifted to the right of the result on *X*. Therefore we can conclude that in this example the addition of privileged information by means of $X + X^*$ provides an inferior solution.

*3.4.2. Clustering by Fusion of X with $X^*$ - ($X = X^*$)*

By using our *aRi-MAX* method, results show a dramatic improvement with regards to the consistency of the solutions. Even if the information in the privileged dataset is obscured by noise, a more consistent solution using the K-Means algorithm was achieved with $\sigma = 0$ for both cases of privileged information. To evaluate the results statistically we performed the Wilcoxon paired signed-rank test again and summarised the results in Table 2 for the $d = 0.2$ case.

These results show that addition of privileged information to the technical data by $X+X^*$ has negative effect on the capability of the K-Means algorithm. Conversely, the use of the proposed *aRi-MAX* method gives results that surpass other results in consistency and therefore in performance. For the case when $d = 0.5$, the statistical results confirm that the fused privileged data in the original technical space provides the K-Means algorithm enough information for achieving a very good solution in the majority of performed runs.

## 4. Can Privileged Information Improve Clustering?

Having established experimental evidence which shows that the use of privileged information in a unique way can be beneficial to the task of clustering, we are now interested in exploring the possibility of using knowledge encoded within privileged information for *improved* clustering. As mentioned in previous section, this task is problematic mainly due to the fact that as we deal with unsupervised learning, we are unable to discriminate between which groups or classes a feature vector belongs to. Thus in turn we are unable to distinguish which group or cluster a piece of information encoded in $X^*$ should affect. Unlike in supervised learning, where privileged information is used for making decisions about the slack variable for the creation of the decision boundary in the SVM algorithm, we can only deal with relative relations between groups of data and their properties across the two hypothetical spaces *X* and $X^*$. There are two main methods which can deal with information, independent of class assignments [33]. The Bayesian approach and the information theoretic approach. Both of these are linked as they essentially deal with probabilities, however their approach to information is slightly different. The Bayesian approach deals with the likelihood of events occurring or their frequencies, possibly given some knowledge, whereas information theoretic approach deals with the uncertainty of events and the amount of information that is encoded within a sequence that describes such events. In our work we are interested in the information theoretic approach.

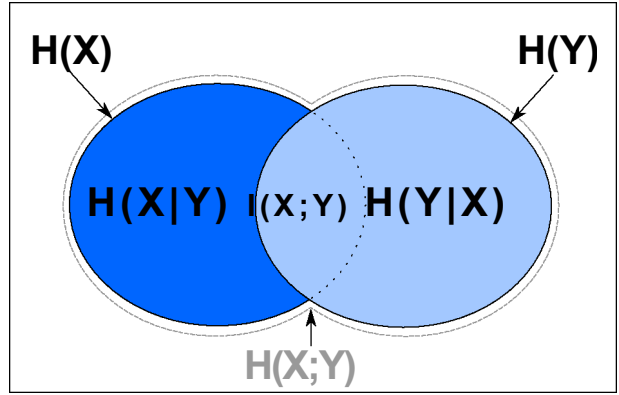

Figure 7: Depiction of the relationship between various information theoretic concepts. H(X) and H(Y) denote entropy of variables X and Y respectively. H(X|Y) denotes the conditional entropy of X given Y. H(X;Y) is the joint entropy of variables X and Y and I(X;Y) is the mutual information of the aforementioned variables.

*4.1. Information Theoretic Approach*

Information theory deals with the quantification of information. It was originally inspired by a subfield of physics called thermodynamics, where the measure of uncertainty about a system is measured using a concept called *entropy*. Entropy in this case is the amount of information that is necessary for the description of the state of the system. At a lower level, entropy is the inclination of molecules to disperse randomly due to thermal motion. The more dispersion, the larger the entropy [15, 12]. Ludwig Boltzman devised a probabilistic interpretation of such thermodynamic entropy,

$$S = k \log W \qquad (3)$$

where the entropy of the system $S$ is the logarithmic probability of its occurrence, up to some scalar factor, k, the Boltzmann constant. Subsequently it was observed that the properties of entropy can be found across many different fields of science. Claude E. Shannon introduced in 1948 the concept of entropy for the purpose of communication over a noisy channel [31], which started the field of information theory. In our work we are dealing with Shannon's concept of entropy and its subsequent variations and derivations.

From the information theoretic point of view entropy is defined as follows,

$$H(X) = - \sum_{i=1}^{n} p(x_i) \log_b p(x_i) \qquad (4)$$

where $H(X)$ is the entropy of a random variable $X$ and $p(x)$ is the probability mass function of an instance $x$ of $X$. Georgii [15] states that entropy should not be considered a subjective quantity. Entropy is essentially a measure of the complexity inherent in $X$, which describes an observer's uncertainty about the variable $X$. Now considering we have two random variables and we would like to find out the information that the two variables share. In other words we would like to know the mutual dependence of variable $X$ on variable $Y$. For this purpose the following calculation, called *Mutual Information* exists,

$$I(X; Y) = \sum_{y \in Y} \sum_{x \in X} p(x,y) \log_b \frac{p(x,y)}{p1(x)p2(y)} \qquad (5)$$

where $p(x, y)$ is the joint probability of variables $x$ and $y$ and $p(x)$ and $p(y)$ are the probability mass functions of variables $x$ and $y$ respectively. The relationship between these information theoretic measure can be seen in Figure 7.
Strehl and Ghosh [33] applied the concept of Mutual Information to the comparison of clusterings, where a modi-fied version of the Mutual Information concept, now normalised to return values in the range [0, 1], has been proposed,

$$NMI(X, Y) = \frac{I(X; Y)}{\sqrt{H(X)H(Y)}} \quad (6)$$

In this equation $I(X; Y)$ denotes the Mutual Information between variables $X$ and $Y$. $H(X)$ and $H(Y)$ denote the entropy of variables $X$ and $Y$ respectively.

The above information theoretic method provides a way to measure and compare the levels of information across different variables. This provides a tool that allows for an insight into combining separate sets of data and extracting the necessary segments of such data that could contribute to an improved solution, without the need for explicit class labels.

*4.2. Dot Product Ratio Measure*

To account for information encoded in the privileged dataset, a method has to be devised that is able to discern which group or cluster should be affected by the information in $X^*$. As our data does not have labels, such processing can only take place according to similarity or consensus between solutions of two or more methods that provide us with their understanding of the underlying data structure. In section 3.3, we have proposed a method to evaluate a number of solutions of the K-Means algorithm, run on the two disparate datasets. In this case one dataset, the technical data, is considered as the dominant set. The solution found for the privileged dataset is only used to find an agreement in order to select the most similar solution for the *dominant* set. An apparent issue arises with this method however. In cases where both solutions end in local minima, this might result in an agreement between the two solutions that is stronger than a more accurate solution, which does not agree strongly across the two datasets. The use of methods which do not have issues with local minima partially resolves this issue. Also our *aRi-MAX* solution only selects the best solution from the technical dataset. It does not provide for an improvement, based on the privileged information. Thus if data in $X$ holds information that is not encoded in $X$, we cannot exploit this. For this reason we must go beyond consensus and attempt to use data in $X$ to amend the solution of $X$.

The first problem with attempting this is how a solution can be affected in a general enough manner to make our method applicable across a wide variety of clustering methods. If we adapt the approach taken by Vapnik [37], we are posed with a number of difficulties. First of all, as mentioned above, we do not have labels. Secondly in order to affect a decision boundary in a similar manner to SVM+, we need to be able to empirically assess our level of correctness of assigning a data item to a specific cluster. Without labels this is impossible. However even if this was possible, for some algorithms this does not pose a reasonable solution. When considering our artificial dataset, the amendment of the cluster centres found by the K-Means algorithm shifts the decision boundary toward a solution which in this case might be correct, however this only works when our data is linearly separable. Any movement of the cluster centres affects the decision boundary, which when moved, might lower the quality of the final solution overall. Thus rather than directly affecting a decision boundary or a cluster centre, we propose to generate additional dimensions or attributes, which encode the best possible solution of a clustering algorithm on the privileged data. In this manner we can try to avoid the issue of linear separability as a non-linear problem might become linearly separable in some arbitrary high dimensional space [30].

The issue of which cluster should be affected by information in $X$ is still pertinent however. In order to deal with this we propose to use a *dot-product* based ratio measure which evaluates whether an item in $X$ is more likely to have been correctly clustered than a related item in $X^*$. Our approach is depicted in Figure 8, which shows how a feature vector $x_i$ is being evaluated against the clustering on $X$ and $X$ and the relative ratio of the location of this vector with respect to its assigned cluster centres is used to determine whether the data item is classified correctly or whether its class should be changed and re-evaluated.

More mathematically we are concerned with calculating the distance of the point $x_i$ projected onto a plane connecting the two cluster centres $C_1$ and $C_2$. To perform such operation we use the dot product,

$$X \cdot Y = |X||Y|\cos\theta \quad (7)$$

where $X$ denotes a vector in the mathematical sense, originating at $C_2$ and connecting $x_i$. $Y$ is a vector connecting the two cluster centres, see Figure 8. $\cos\theta$ is the angle between these two vectors, however as we do not know the value of this angle, we use the dot product form which calculates the projection $Z$ using the known lengths of $X$ and $Y$ as follows,

$$Z = \frac{X \bullet Y}{Y \bullet Y} Y \qquad (8)$$

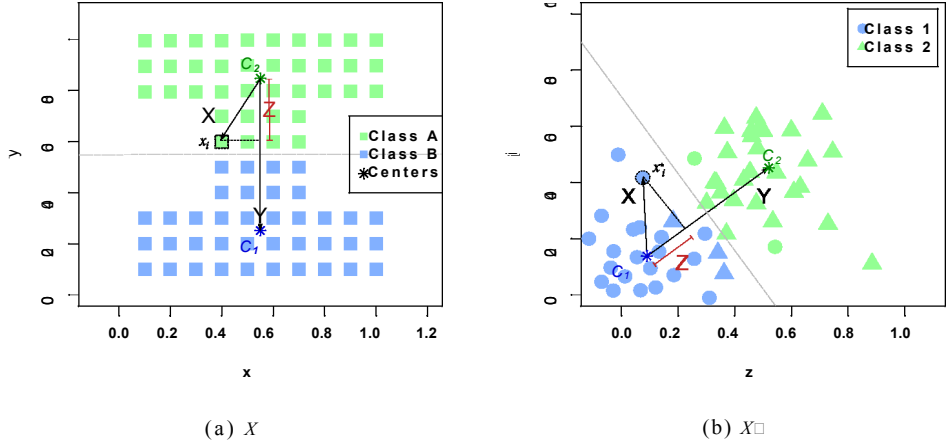

(a) $X$                                (b) $X^\star$

Figure 8: Depiction of the dot product ratio measure used to determine whether an assignment of an input feature vector is more likely to belong to a cluster as determined by clustering of $X$ (a) or to a cluster as determined by clustering of $X^\star$ (b). In our approach this decision depends on the comparison of relative ratios of distance from the vector's assigned cluster centre for each clustering. In this figure $Y$ denotes a vector connecting the two cluster centres, $X$ denotes a vector connecting input $x_i$ with the closest cluster centre and $Z$ denotes the projection of $X$ onto $Y$.

To calculate the necessary ratio which allows us to evaluate how correct a clustering solution might be for a particular point $x$, we perform the following calculation,

$$R_d(x) = \frac{\|Z - C_k\|}{\|Z - C_{k+1}\|} \qquad (9)$$

where $R_d(x)$ denotes the function that calculates the distance ratio on $x$, $Z$ denotes the projection of the currently examined input $x_i$ onto the vector $X$ and $C_k$ and $C_{k+1}$ are the assigned cluster centre and the closest subsequent cluster centre respectively. In cases where we only have two clusters, $C_{k+1}$ denotes the cluster centre to which $x_i$ has not been assigned, in other words the opposing cluster centre. In cases where $k > 2$ the cluster centre $C_{k+1}$ is the closest cluster centre according to some metric, such as the Euclidean distance or the topographic distance in the SOM algorithm. In our experiments we only deal with $k = 2$ to simplify our analysis. The symbol $\|\cdot\|$ denotes the Euclidean norm and in our calculation the distance between the projected location $Z$ from its assigned cluster centre as well as from the opposing cluster centre. The projection of point $x_i$ is calculated according to Equation 8 as follows,

$$Z = \frac{(x_i - C_k) \bullet (C_{k+1} - C_k)}{(C_{k+1} - C_k) \bullet (C_{k+1} - C_k)}(C_{k+1} - C_k) \qquad (10)$$

where $x_i$ denotes our input vector of interest and $C_k$ and $C_{k+1}$ again the cluster centres computed by a clustering algorithm. To put this calculation into perspective the pseudo-code for our method, called *P-Dot*, is shown in Algorithm 2 and explained in detail in the next paragraph.

*4.3. Privileged Information Dot Product Consensus:* P-Dot

Having established a method for evaluating whether an input from $X$ is more likely to have been correctly clustered than the input from $X^\star$ or vice versa, we now describe our algorithm that combines this method for the benefit of improved clustering. Pseudo-code 2 highlights each step of our proposed *P-Dot* algorithm.

**Algorithm 2:** P-Dot Algorithm

**Input**: Technical Data: *X*
**Input**: Privileged Data: *X*\**
**Output**: Clustering of *X*

**1** initialization;
**2 foreach** *i* in *iter* **do**  /* Consensus Step */
**3**   $Ctech_i \leftarrow$ clust (*X*);
**4**   $Cpriv_i \leftarrow$ clust (*X*\**);
**5 end**
**6 foreach** *i* in $Ctech_i$ **do**
**7**   $Cbest_p \leftarrow$ max (NMI ($Ctech_i$;$Cpriv_i$)) ;  /* Mutual Information */
**8 end**
**9** $Ctech_{best} \leftarrow$ which (max ($Cbest_p$));
**10** $Cpriv_{best} \leftarrow$ max ($Cbest_p$);
**11 if** *NMI ($Ctech_{best}$;$Cpriv_{best}$) < min (H ($Ctech_{best}$);H ($Cpriv_{best}$))* **then**
**12**   **if** differences > matches **then**  /* Fusion Step */
**13**     $x_i \leftarrow$ matches ($Ctech_{best}$;$Cpriv_{best}$);
**14**   **else**
**15**     $x_i \leftarrow$ differences ($Ctech_{best}$; $Cpriv_{best}$);
**16**   **end**
**17**   **foreach** *i* in $x_i$ **do**
**18**     **if** $R_d(Ctech_{best} [x_i])$ > $R_d(Cpriv_{best} [x_i])$ **then**  /* Ratio Measure */
**19**       swap.cluster.assignment($Ctech_{best} [x_i]$);
**20**     **end**
**21**   **end**
**22**   $X_{new} \leftarrow$ bind (*X*, $Ctech_{best}$ [A], $Ctech_{best}$ [B]);
**23**   $C_{final} \leftarrow$ clust ($X_{new}$);
**24 end**

In Algorithm 2 lines 2 to 10 highlight an amended version of our consensus method described in section 3.3. Instead of using the Adjusted Rand Index method, we employ here the Normalised Mutual Information (Equation 6) to find two most similar clusterings amongst a set of solutions. Strehl and Ghosh [33] have shown that this measure is a suitable way of comparing clusterings with some advantages over the Adjusted Rand Index method, such as the absence of strong assumptions on the underlying distribution. Steps 11 to 24 outline each step of the *P-Dot* method. Initially the results of the best clustering solutions are evaluated for similarities at steps 12-16. If the two clusterings are identical, then no improvement can be obtained. If there exist differences between the solutions, we determine whether there are more differences than similarities between the two solutions (lines 12-14) or vice versa (lines 1516). This is due to the fact that explicit cluster labels are meaningless and thus cluster A on $X$ can be called cluster B in $X^*$. For this reason we only need to deal with the smaller set of values, whether they are matching labels or dissimilar labels. Then for each match or mismatch (line 17), the dot-product ratio $R_d(x)$ is calculated (line 18) for both inputs from $X$ and $X^*$, to determine which of the two solutions is more likely to be correct. If the solution on space $X$ is more likely to be correct, the label of the corresponding item in the solution of $X^*$ is inverted (for $k = 2$), at step 19. Once this procedure is completed, $k$ new attributes are created, one per each cluster found (line 22). These attributes are filled with maximum normalised values for data items that belong to that particular cluster according to the labels as decided in the previous step. Eventually a new dataset $X_{new}$ is generated. This data is clustered again and its solution is the final solution of our method (line 23). What this approach achieves is in effect taking into account data streams separately and treating them uniquely. This allows for data that might normally be deemed irrelevant due to swamping, to become equally important in the final processing of the dataset.

*4.4. Experimental Evaluation*

To evaluate our proposed method empirically we employ our artificial dataset from paragraph 3.4. The K-Means clustering algorithm is used within the *P-Dot* method and compared against our previous *aRi-MAX* method as well as a set of four established clustering methods. These methods are the Expectation Maximization algorithm [11], which is a probabilistic algorithm that fits Gaussian models onto the dataset, Spectral Clustering [32, 22], which transforms the original data into another space within which the actual clustering is performed. Also two versions of the SOM [18] algorithm are employed. One method that treats the SOM as a clustering algorithm assigning one node per cluster, similarly to K-Means and another method which specifies a larger number of nodes that are shared between clusters present within the dataset. These nodes are subsequently clustered using the K-Means algorithm in the same way as was performed by Vesanto and Alhoniemi in [40]. The mixture of $X + X^*$, combined in the same feature space is used as the dataset on which the comparison techniques are employed.

First our *P-Dot* method is tested on the artificial dataset with point-wise (perfect) privileged information. In both cases ($d = 0.2$ and $\check{d} = 0.5$) the performance of our new method is superior to the *aRi-MAX* method. The *P-Dot* method achieves almost 100% accuracy in both cases, compared to 62% performance of the *aRi-MAX* method. Subsequently when the more complex Gaussian based privileged information is tested, again our new method achieves very good results in both situations. An average concordance of 85% and 95% for the $d = 0.2$ and $\check{d} = 0.5$ cases are achieved, respectively. This is an improvement of 23% and 33%, respectively, over the *aRi-MAX* method.

When compared to other existing techniques, such as the SOM, Spectral Clustering and Expectation Maximization algorithms, we can see that our method is still superior in both cases where privileged information is more complicated ($\check{d} = 0.2$). Box plots for these results can be seen in Figures 9(a) and 9(b) respectively.

Table 3 highlights a summary of statistics for all of the tested algorithms on the Gaussian privileged dataset with $\check{d} = 0.2$. It is apparent that our *P-Dot* method outperforms all others in all aspects except possibly stability. However the standard deviation of the results for our method is still very low. Best results in this table have been underlined for the purpose of clarity.

**5. P-Dot in the real world**

To fully assess the usefulness of our *P-Dot* method, we need to apply it to and analyse it within a much more complex scenario. The existence of privileged information in existing datasets is not prevalent due to the novel nature of the paradigm proposed by Vapnik. A dataset created for the evaluation of such type of learning has however been created in [38]. We will use this dataset for our analysis and for comparison with existing clustering techniques.

**Table 3: Statistical information on the performance of the *P-Dot* method compared to various clustering techniques applied to $X + \sqrt{X^*}$ using the Gaussian based privileged information with $d = 0.2$ (Figure 6a). The values show the normalized mutual information between clusters found using a given method and the true class labels across 100 runs. 1 denotes perfect match and 0 denotes no matches at all. Best results have been underlined. *P-Dot* performs best overall. SOM2K denotes the combination of SOM and K-Means according to [40].**

|                      | Min     | Max    | Mean   | Median | St.Dev. |
|----------------------|---------|--------|--------|--------|---------|
| K-Means ($X$)        | -0.0129 | 0.6184 | 0.4719 | 0.6184 | 0.2623  |
| K-Means ($X + X^*$)  | -0.0129 | 0.6184 | 0.5373 | 0.6184 | 0.2109  |
| aRi-MAX              | 0.6184  | 0.6184 | 0.6184 | 0.6184 | 0.0000  |
| *P-Dot*              | 0.8462  | 0.8960 | 0.8472 | 0.8462 | 0.0070  |
| EM                   | 0.7979  | 0.7979 | 0.7979 | 0.7979 | 0.0000  |
| Spectral             | 0.6184  | 0.7979 | 0.6902 | 0.6184 | 0.0884  |
| SOM                  | 0.6184  | 0.6184 | 0.6184 | 0.6184 | 0.0000  |
| SOM2K                | -0.0162 | 0.7054 | 0.4395 | 0.5770 | 0.2629  |

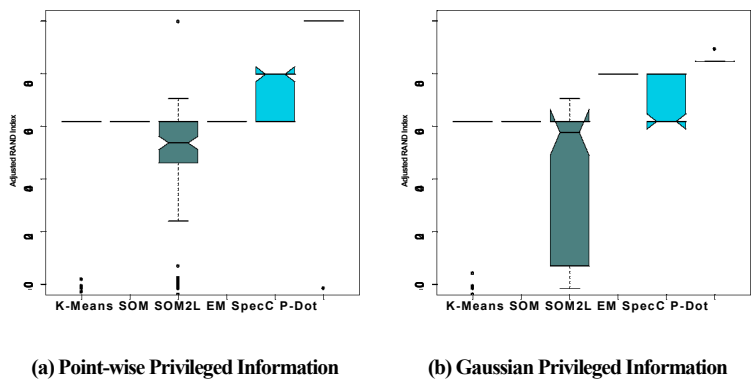

(a) Point-wise Privileged Information     (b) Gaussian Privileged Information

**Figure 9: Box plots comparing the diverse clustering techniques tested against our proposed *P-Dot* method. These plots provide a clear overview of the results, confirming our *P-Dot* method performing best in both cases.**

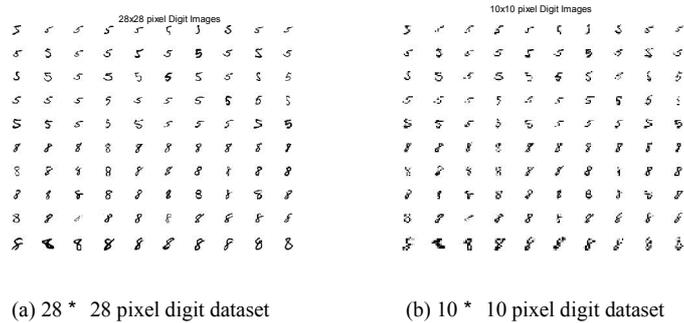

(a) 28 * 28 pixel digit dataset  (b) 10 * 10 pixel digit dataset

Figure 10: Subset of the MNIST digit dataset comprising of 100 digits at two different resolutions, providing different levels of information. The lower the resolution, the more information is potentially lost. The dataset comprises of two classes of digits, the numbers 5 and 8 and variations thereof. There are 50 examples of each class.

*5.1. The Task of Digit Recognition*

The task of digit recognition is one of the most frequent pattern recognition tasks. Numerous research papers and results have been generated on this topic, particularly within the supervised learning community. The MNIST database of handwritten digits [19] is an eminent source of such research[1]. Vapnik [38] used a subset of the MNIST dataset comprising of 100 digits from two different groups. The two groups are 50 variations of the digit 5 and 50 variations of the digit 8. Each digit is originally a 28*28 pixel gray-scale image, seen in Figure 10(a). Thus an image can be thought of as a 784-dimensional feature vector with values in the range $[0 - 255]$. To make the task of classification slightly more difficult, Vapnik created a second dataset, based on scaled down versions of the original images. Thus a set of 100 gray-scale images at a resolution of 10*10 pixels has been created, seen in Figure 10(b). This dataset can again be thought of as a 100-dimensional dataset with a range $[0 - 255]$.

*5.2. Privileged Information*

To explore the notion of learning using privileged information, such type of additional information had to be created. In order to do this Vapnik et al. [38] created a set of poetic descriptions with the help of language experts. By poetic description we mean a description of what the expert saw and interpreted using his own words in the form of a poem. An example of such a description follows:

> **Poetic Description:** *"Item.4 - A two-part contradictory creature. One part is rounded, the other is a bit angular. The bottom is on the earth (steadier than the first three figures). Open and free for the wind. It is slightly slanted to the right. The upper part is broken. It has a small gap. Insignificant. The lower round part has no "hill". The man throws a stick. He goes headfirst. He looks ahead - where is the stick? He is attacking somebody or training. The wriggling snake is beating its tail. The unpleasant movement. Not so nice. A bit irregular. Not so clear. A rope. Asymmetrical. No curlings of the ends."*

In addition to this description a separate interpretation of the digits has also been created. Rather than a subjective description of the emotive impact of each digit on the observer using arbitrary words, the second set of privileged information has been created using words of opposing meanings. For this purpose this set of privileged data is termed Ying Yang[2]. An example of this type of data follows:

> **Ying Yang:** *"Item.3 - Very sociable creature between 30 and 40. Everything is normative and rightful. Masculine firmness in everything. May be dull and uninventive but good-natured. Seeking nothing. It is still water but useful with all the usual advantages. Take your time - you can live nearby and be happy. Nothing mysterious about it. You see the bottom, the water is so transparent. Simplicity may be its strong side."*

---

[1]The MNIST dataset can be downloaded from http://yann.lecun.com/exdb/mnist/. This website also contains a summary of the best published classification results of various supervised algorithms using this data

[2]An ancient Chinese philosophy of the interconnection and interdependency of opposing forces within the natural world

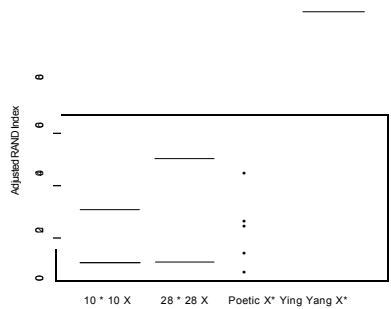

Figure 11: Analysis of the K-Means algorithm on $X$ itself as well as on both sets of privileged data $X^*$ on their own. It is apparent that the higher resolution data contains more information, useful for better clustering. The Poetic $X^*$ data is not a useful resource of information on its own when clustered using K-Means. The opposite is true for the Ying Yang $X^*$, which allows for correct identification of the two true clusters on its own.

To make these additional sets of data usable in computation, the text has been analysed and a set of keywords that occur across the dataset have been extracted. An example of such a keyword is "Upright", which captures whether a digit is slanted or not, or "Gaps", which measures the amount of gaps a digit has. For each keyword a scale of the term's possible range of meaning has been proposed. In our example keywords "Upright" has a binary value and "Gaps" has a scale from 0 to 5. Finally a 21-dimensional vector has been created for each digit encoding the poetic description of each digit and a 31-dimensional vector has been created for the Ying Yang dataset[3].

*5.3. Clustering Performance*

To evaluate the capability of our proposed *P-Dot* method, we have used the above described dataset and performed a number of experiments. We have run each algorithm 100 times to ensure better consistency for the comparison of results. First we used the K-Means algorithm and clustered each individual dataset on its own, to understand how well each dataset can segment the data into the correct categories. The result of this clustering, Figure 11, provides an insight into the approximate level of information encoded within the datasets that can be revealed using the K-Means method. The higher resolution version of the data gives a slightly better insight into the data's underlying structure and thus the K-Means performs better in this scenario, by approximately 3%. When looking at the privileged information, we observe that there is a dramatic difference between the two sets of data. The poetic description of each digit reveals very little on average, with a hint that in cases when the K-Means algorithm starts with a good set of initial values it is possible to achieve clustering that has an agreement of at maximum 38%. On the other hand when looking at the Ying Yang privileged information we can see that the dataset can be categorised into the two required groups with 100% accuracy, consistently, using just the privileged information. From a logical point of view this difference between the privileged information makes sense as the Ying Yang descriptions seek to evaluate visual features that are contradictory and thus aim at describing each digit at extreme ends of the description scale. For our analysis we are mainly interested in exploring the benefit of privileged information with a discrimination ability such as the poetic dataset as we believe such type of additional information, which is insufficient on its own to solve the task, is more likely to occur in many real-world situations.

Subsequently an evaluation of the clustering performance of the K-Means dataset on the fusion $X + X^*$ is required to be able to assess if such a combination provides good enough results on its own. The data has been standardised using the min-max method [27, 16] in order to ensure that no swamping of attributes can occur. The result of the K-Means clustering on the normalised version of the space $X$ can be seen in Figure 12. The fusion of the technical and the privileged data resulted in *decreasing* the quality of the K-Means solution in both cases by up to 6%. Thus we can assume that without any further processing the additional information in the original feature space only hinders our search for a good solution. This highlights the necessity of processing such data in a unique way that allows for the extraction of information from $X^\iota$ that can improve the final cluster, but not *degrade* it.

---

[3]The reader is referred to *http://www.nec-labs.com/research/machine/ml website/department/ software/learning-with-teacher/* where a detailed description of the dataset exists.

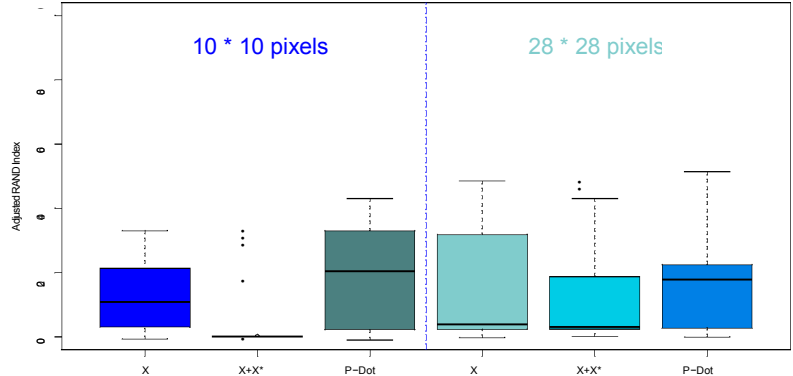

Figure 12: Comparison of results of the K-Means algorithm and the *P-Dot* method on the digit dataset. Results for individual clusterings of *X* for both versions of the normalised datasets as well as the clustering on the fused $X + X^\iota$ data using K-Means are shown. The $X + X^\iota$ fusion clearly degrades the performance of the K-Means algorithm. Results of the *P-Dot* method using Poetic $X^\iota$ on the other hand show an improvement over K-Means in both situations. An improvement over *X* and more dramatically over $X + X^\iota$ is apparent.

Knowing that privileged information fused within the feature space of the technical data does not bring any benefit to cluster analysis, our *P-Dot* method is employed to assess the benefit of data from $X^\iota$ combined in our unique way. For the lower resolution dataset we observe that our method provides an improvement over both, *X* only, as well as the fused $X + X^*$, seen in Figure 12. This is the case for both the mean performance across the 100 runs, where an improvement of 5% with respect to clustering on *X* is observed and 11% with respect to $X + X^*$. The best possible attainable solution also improved in this case. In the case of the higher resolution version, there is an improvement of 1% with respect to *X* and 6% with respect to $X + X^*$. This is a promising result as we only want to improve upon a solution in case the privileged data can provide such benefit. When such improvement is unattainable from the additional dataset, we do not want to negatively affect our existing solution. Wilcoxon signed-rank statistical test was performed, confirming these results. Very small p-values for the two-sided tests reject the hypothesis that the two results are the same. Analysis of the alternative hypotheses help us clearly identify that our *P-Dot* method has better results in both cases.

### 5.4. The Importance of Dimensionality Reduction in P-Dot

When considering how the *P-Dot* algorithm works, one might realise that the higher the dimensionality of the data, the smaller the influence of combining the privileged data with the technical data on the final outcome. The way the *P-Dot* algorithm amends the solution based on privileged data is based on adding additional dimensions which encode the consensus view of the clusterings of the two separate hypothetical spaces. Thus in the case of the higher resolution digit dataset, the *P-Dot* algorithm essentially adds two more dimensions to the existing 784 dimensions. For this reason even when all attributes are normalised, the influence of the two additional attributes is negligible. For this reason we believe that the consensus attributes generated by our method need to be evaluated at a comparable level to the rest of the data. One way of achieving this is to perform dimensionality reduction. One of the most frequently used methods for such a task is the Principal Component Analysis (PCA) method developed by Karl Pearson [24]. The PCA is a non-parametric transformation method that transforms data to a new coordinate system, where data with the largest variance are projected on the first coordinate. Such coordinate is called a principal component. Data of subsequent smaller variances are projected on the second and then subsequent coordinates. Eventually the whole dataset is transformed in a way which emphasizes uncorrelated variables where the first variable captures the majority of variance of the whole dataset. For this reason, when the PCA is used for dimensionality reduction, traditionally the first two principal components are used in analysis as they generally capture the majority of the data's underlying structure. In our next experiment we have employed the PCA technique on the technical data in order to reduce the dimensionality from 784 dimensions to only two. Our experiments have shown that such a transformation does not hinder greatly the clustering capability of the K-Means algorithm with respect to clustering on $X$. On the contrary the reduction has selected the most relevant features of the $X + X^*$ data which more clearly reveal the underlying structure and thus improve performance with respect to the clustering on $X + X^*$. When the PCA is used on $X$ as part of our *P-Dot* algorithm, there is no apparent difference for the lower resolution dataset with respect to the *P-Dot* method on normalised $X$. On the other hand the higher resolution dataset benefits significantly from this reduction and the *P-Dot* method can have greater influence on the final clustering result. This results is a further improvement of approximately 5% on top of the *P-Dot* method on normalised $X$. Statistical tests were performed on all results to confirm their validity.

We believe that there is a fundamental issue that needs to be addressed that is missing from many machine learning experiments. The fact that our *P-Dot* method could improve upon the solution of a traditional method on the same dataset, using only a form of an agreement and subsequent encoding of such an agreement within a dataset that is re-evaluated, hints at the possible importance of not only the notion of privileged information, but more interestingly on the importance of different generators of data. Privileged information can be thought of as simply an additional set of attributes that describe a given problem. Our analysis above however showed that when adding such data to existing technical data in the traditional way, no benefit is apparent. What is necessary for this data to become useful is, we believe, the relative importance of the information encoded within, to be treated with equal weight as data from any other source, independent of dimensionality. In other words, one attribute from a separate source or generator of data should be treated equally to any number of attributes from another, but single source.

*5.5. Comparison with Other Clustering Techniques*

To contrast our *P-Dot* method with other existing clustering techniques, we have applied the same set of clustering algorithms used in section 4.4 to the digit datasets. An overview of the results obtained for the digit dataset can be seen in Figures 13 and 14. All datasets except our *P-Dot* method were run on the normalised fused version of the digit data ($X + X^*$). From the results we can observe that all algorithms, except Spectral Clustering and our *P-Dot* method, perform very badly in the 10x10 case, shown in Figure 13(a). We believe that this is due to the effect of the fusion of the two datasets. By combining the two in the original feature space, the task of finding the correct underlying structure within the data becomes more difficult. A possible reason for Spectral Clustering performing better than other techniques is that this method is a data transformation method that transforms data to a new metric space where distances are based on flow rather than just one metric distance, upon which a clustering is performed. In this new metric space non-linear problems can become linearly separable and thus the Spectral Clustering algorithm is able to perform well even with the possible introduction of noise or data that make the problem more complex. Our *P-Dot* method performs best, in terms of the mean agreement between the method's clustering and the true clustering, with a concordance of approximately 19%, followed by Spectral Clustering at 14% in both datasets.

When we subject the dataset to PCA, the results for majority of the tested algorithms change, Figure 14. This suggests that the variance of the underlying data is an important property that is required for correct clustering of the data. The result of our method does not show any significant difference from the untreated dataset, however the *P-Dot*'s result is still the highest amongst all the tested algorithms for the 10x10 dataset, see Figure 14(a). Both the EM algorithm and Spectral Clustering benefit from the PCA treatment and their results improve dramatically. The mixture of SOM with K-Means in a second layer also performs substantially better and is almost on par with our approach. Statistical overview of results for all tested algorithms for this scenario are shown in Table 4.

The performance of the tested algorithms on the high resolution dataset can be seen in Figures 13(b) and 14(b). It is apparent that the performance of all the algorithms on the 28x28 dataset that has only been normalised is still very low. In this case, on average, our *P-Dot* method still performs the best. However once the dataset undergoes dimensionality reduction, the EM algorithm is able to determine the structure of the underlying data very well and achieves a very good result of almost 50% agreement between the solutions and the correct clustering. Thus in this scenario our method comes second, only after the EM method.

It is widely accepted that different clustering techniques are applicable to different problems. Some methods are simply more suitable to specific types of data. This has been highlighted by our results in this case. Taking this into account, we can apply our *P-Dot* method while using a different underlying method than the K-Means algorithm. For this purpose we have used the EM algorithm as the last step of the *P-Dot* process (step 23 in Algorithm 2). The results of this mixture are contrasted with the EM algorithm run on the fusion of $X + X^*$. By combining two different clustering methods within our *P-Dot* approach we are able to achieve even better results for the 10x10 dataset than using any other tested method. An average agreement of 31% is achieved in this case, confirming our intuitions.

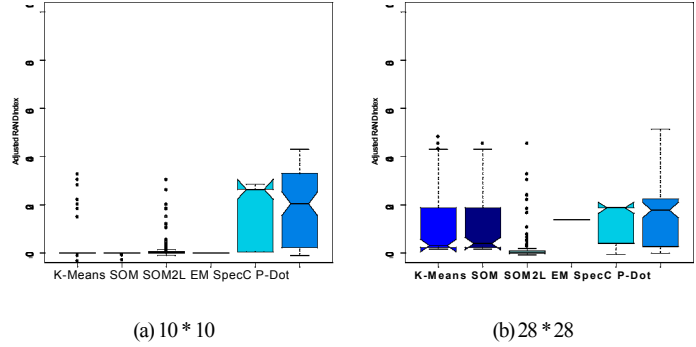

(a) 10 * 10

(b) 28 * 28

Figure 13: Performance of diverse clustering algorithms on the digit dataset, normalised and fused by $X + X^*$. The box plots of these results confirm a good performance of our *P-Dot* method in both scenarios. Spectral Clustering also performs well in comparison to other methods and the EM algorithm performs well in the 28x28 case.

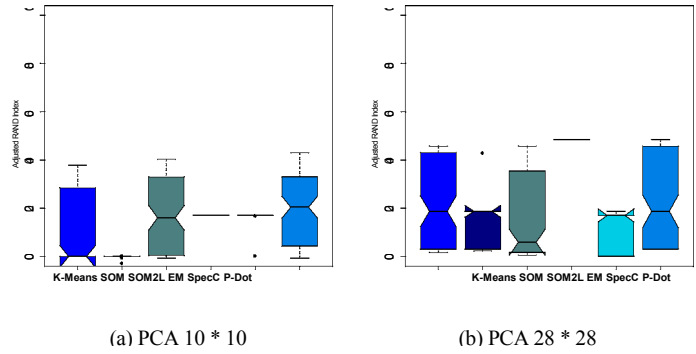

(a) PCA 10 * 10

(b) PCA 28 * 28

Figure 14: Performance of diverse clustering algorithms on the digit dataset, normalised and fused by $X + X^*$, then subject to PCA. In this scenario the EM algorithm performs best for the 28x28 dataset with our *P-Dot* method coming second. In the lower resolution dataset, our method has a slight edge over other methods, however both Spectral Clustering and the EM algorithm have an advantage in terms of result stability.

Table 4: Statistical results of the clustering algorithm comparison on the 10x10 dataset subjected to PCA. Values show the normalized mutual information between clusters found by a given algorithm and the true class labels across the 100 runs performed. Best results are highlighted for convenience. In this case again our *P-Dot* method performs best with *P-DotEM* achieving the most consistent good result. SOM2K denotes the combination of SOM and K-Means according to [40].

| Algorithm | Min | Max | Mean | Median | St.Dev. |
|---|---|---|---|---|---|
| K-Means ($X + X^*$) | 0.0002285 | 0.3782 | 0.1229 | 0.0002285 | 0.1482 |
| SOM | 0.0000 | 0.004603 | 0.0000 | 0.0002285 | 0.001893 |
| SOM2K | 0.0000 | 0.4036 | 0.1627 | 0.1606 | 0.1492 |
| EM | 0.1708 | 0.1708 | 0.1708 | 0.1708 | 0.0000 |
| Spectral | 0.004489 | 0.1708 | 0.1658 | 0.1708 | 0.02852 |
| *P-Dot* | 0.0000 | 0.4299 | 0.1893* | 0.2047* | 0.1329 |
| *P-DotEM* | 0.3077 | 0.3077 | 0.3077 | 0.3077 | 0.0000 |

## 6. Conclusions and Future Work

In this work we have investigated the importance and incorporation of privileged information in cluster analysis. In the supervised setting Vapnik and colleagues have shown in the past that a new paradigm called "Learning Using Privileged Information" provides a novel approach for learning using data from disparate hypothetical spaces that surpasses traditional supervised techniques. In our work we translate the notion of privileged information to the unsupervised setting in order to improve clustering performance. This allows us to use disparate types of data from many existing machine learning problems in a way that enhances the clustering process, while avoiding the degradation of the quality of clusters when such additional data is incorporated.

First we have highlighted and empirically demonstrated the difference between technical and privileged information. Our analysis confirms that there is a benefit in using information from different hypothetical spaces in a way that is different from simply combining the data in the technical feature space. With the help of an artificial dataset we showed that we can improve the stability of the K-Means algorithm by employing an adjusted Rand index consensus method, we termed *aRi-MAX*, which selects a clustering of the technical data according to the best matching clustering of the privileged data. This form of consensus tends to select the best possible solution of the technical data, as can be achieved with the underlying clustering technique. The drawback of this approach is the increased computational complexity due to the repetitive nature of the consensus selection method and the clustering performance which is limited by the underlying clustering algorithm. One benefit is the possibility of the use of a variety of clustering algorithms with the *aRi-MAX* method and thus the most suitable combination of algorithms can be chosen, depending on the problem in question.

Another positive and important feature is the fact that the two datasets $X$ and $X^*$, which possibly originate from different generators, are treated equally in terms of processing. We believe that this is a very important requirement for correct clustering of any data that includes signals from more than one generator. Even though the separate generators of data are treated equally in our method, only original data is used for final partitioning.

Aiming towards improved clustering performance using privileged data, we exploit privileged information as part of the clustering process itself. One major issue of such task is the lack of knowledge of classes. In the supervised setting this information is available for the learning machine and thus can be exploited and combined with the information encoded in the privileged data stream. In the unsupervised setting we do not know which clusters, classes or groups of data should be affected using our additional knowledge. Thus in order to answer the question whether privileged information can be used to improve clustering performance, we have proposed a novel cluster fusion algorithm based on information theory and the dot-product, called *P-Dot*. This algorithm, when tested on our artificial dataset, has shown very encouraging results, confirming our hypothesis that privileged information can enhance clustering in a more substantial manner than by simply appending this type of information at the end of the original technical data. When comparing our method to other clustering approaches we can still see a considerable improvement over these methods.

By testing our proposed *P-Dot* algorithm on a real world digit dataset, we are able to assess its usefulness for a larger variety of tasks. Our analysis has shown that the proposed method does provide for an improvement in clustering over the classical K-Means algorithm as well as many existing established clustering techniques. Some techniques can achieve very good results in specific scenarios, such as the EM algorithm in the case of the PCA treated fusion of the technical 28x28 dataset with the poetic privileged information. Nevertheless the strength of our approach is the ability to use many existing clustering techniques individually or in combination and thus focus on exploiting the benefits of each technique, where it is needed.

In the future we believe that our method could be improved when considering more than only the solution of a number of clustering algorithms. A form of information theoretic measure which highlights some properties of the underlying data could be used to assess which type of algorithm to use and how much the fusion of the two hypothetical spaces should be encouraged. The creation of a classification model out of the result of the clustering process akin to

# References

[1] S. Ben-David, J. Blitzer, K. Crammer, A. Kulesza, F. Pereira, and J. Vaughan. A theory of learning from different domains. *Machine Learning*, 79(1):151–175, May 2010.
[2] S. Bickel and T. Scheffer. Multi-View Clustering. In *Data Mining, 2004. ICDM '04. Fourth IEEE International Conference on*, volume 0, pages 19–26, Los Alamitos, CA, USA, 2004. IEEE Computer Society.
[3] A. Blum and T. Mitchell. Combining labeled and unlabeled data with co-training. In *Proceedings of the eleventh annual conference on Computational learning theory*, COLT' 98, pages 92–100, New York, NY, USA, 1998. ACM.
[4] P. S. Bradley and U. M. Fayyad. Refining Initial Points for K-Means Clustering. In *ICML '98: Proceedings of the Fifteenth International Conference on Machine Learning*, pages 91–99, San Francisco, CA, USA, 1998. Morgan Kaufmann Publishers Inc.
[5] F. Cai and V. Cherkassky. SVM~ regression and multi-task learning. In *IJCNN'09: Proceedings of the 2009 international joint conference on Neural Networks*, pages 503–509, Piscataway, NJ, USA, 2009. IEEE Press.
[6] N. Cesa-Bianchi, D. Hardoon, and G. Leen. Guest Editorial: Learning from multiple sources. *Machine Learning*, 79(1):1–3, May 2010.
[7] O. Chapelle, B. Schölkopf, and A. Zien, editors. *Semi-Supervised Learning*. Adaptive Computation and Machine Learning. The MIT Press, September 2006.
[8] Y. Chen and Y. Yao. A multiview approach for intelligent data analysis based on data operators. *Information Sciences*, 178(1):1–20, January 2008.
[9] D. Chou, C.-Y. Jhou, and S.-C. Chu. Reversible Watermark for 3D Vertices Based on Data Hiding in Mesh Formation. *International Journal of Innovative Computing, Information and Control*, 5(7):1893–1901, July 2009.
[10] V. de Sa, P. Gallagher, J. Lewis, and V. Malave. Multi-view kernel construction. *Machine Learning*, 79(1):47–71, May 2010.
[11] A. P. Dempster, N. M. Laird, and D. B. Rubin. Maximum Likelihood from Incomplete Data via the EM Algorithm. *Journal of the Royal Statistical Society. Series B (Methodological)*, 39(1):1–38, 1977.
[12] E. Fermi. *Thermodynamics*. Dover Publications, June 1956.
[13] G. Forestier, P. Gançarski, and C. Wemmert. Collaborative clustering with background knowledge. *Data & Knowledge Engineering*, 69(2):211–228, February 2010.
[14] A. Gionis, H. Mannila, and P. Tsaparas. Clustering aggregation. *ACM Trans. Knowl. Discov. Data*, 1(1):4+, March 2007.
[15] A. Greven, G. Keller, and G. Warnecke, editors. *Entropy (Princeton Series in Applied Mathematics)*. Princeton University Press, October 2003.
[16] J. Han, M. Kamber, and J. Pei. *Data Mining: Concepts and Techniques, Second Edition (The Morgan Kaufmann Series in Data Management Systems)*. Morgan Kaufmann, 2 edition, January 2006.
[17] L. Hubert and P. Arabie. Comparing partitions. *Journal of Classification*, 2(1):193–218–218, December 1985.
[18] T. Kohonen. Automatic formation of topological maps of patterns in a self-organizing system. In *Proceedings of the 2nd Scandinavian Conference on Image Analysis*, pages 214–220, Espoo, 1981.
[19] Y. Lecun, L. Bottou, Y. Bengio, and P. Haffner. Gradient-based learning applied to document recognition. *Proceedings of the IEEE*, 86(11):2278–2324, August 2002.
[20] L. Liang, F. Cai, and V. Cherkassky. 2009 Special Issue: Predictive learning with structured (grouped) data. *Neural Netw.*, 22(5-6):766–773, July 2009.
[21] A. Likas, N. Vlassis, and J. J. Verbeek. The global k-means clustering algorithm. *Pattern Recognition*, 36(2):451–461, February 2003.
[22] A. Y. Ng, M. I. Jordan, and Y. Weiss. On Spectral Clustering: Analysis and an algorithm. In *Advances in Neural Information Processing Systems 14*, volume 14, pages 849–856, 2001.
[23] J. M. Peña, J. A. Lozano, and P. Larrañaga. An empirical comparison of four initialization methods for the K-Means algorithm. *Pattern Recognition Letters*, 20(10):1027–1040, October 1999.
[24] K. Pearson. On lines and planes of closest fit to systems of points in space. *Philosophical Magazine*, 2(6):559–572, 1901.
[25] D. Pechyony, R. Izmailov, A. Vashist, and V. Vapnik. SMO-style algorithms for learning using privileged information. In *Proceedings of the 2010 International Conference on Data Mining (DMIN'10)*, 2010.
[26] D. Pechyony and V. Vapnik. On the Theory of Learning with Privileged Information. In *Advances in Neural Information Processing Systems 23*, 2010.
[27] K. L. Priddy and P. E. Keller. *Artificial Neural Networks: An Introduction (SPIE Tutorial Texts in Optical Engineering, Vol. TT68)*. SPIE Publications, illustrated edition edition, 2005.
[28] W. M. Rand. Objective Criteria for the Evaluation of Clustering Methods. *Journal of the American Statistical Association*, 66(336):846–850, 1971.
[29] B. Ribeiro, C. Silva, A. Vieira, A. Gaspar-Cunha, and J. C. das Neves. Financial distress model prediction using SVM~. pages 1–7, July 2010.
[30] B. Schölkopf, K. Tsuda, and J.-P. Vert. *Kernel Methods in Computational Biology*. Computational Molecular Biology. The MIT Press, August 2004.
[31] Shannon and W. Weaver. A mathematical theory of communication. *Bell Syst. Tech. J*, 27:379–423, 1948.
[32] J. Shi and J. Malik. Normalized cuts and image segmentation. *IEEE Transactions on Pattern Analysis and Machine Intelligence*, 22(8):888–905, Aug 2000.
[33] A. Strehl and J. Ghosh. Cluster ensembles — a knowledge reuse framework for combining multiple partitions. *Journal of Machine Learning Research*, 3:583–617, March 2002.
[34] W.-L. Tai and C.-C. Chang. Data Hiding Based on VQ Compressed Images Using Hamming Codes and Declustering. *International Journal of Innovative Computing, Information and Control*, 5(7):2043–2052, July 2009.
[35] A. Topchy, A. K. Jain, and W. Punch. Clustering ensembles: models of consensus and weak partitions. *Pattern Analysis and Machine Intelligence, IEEE Transactions on*, 27(12):1866–1881, October 2005.
[36] V. Vapnik. *Estimation of Dependences Based on Empirical Data (Information Science and Statistics)*. Springer, March 2006.

[37] V. Vapnik and A. Vashist. A new learning paradigm: Learning using privileged information. *Neural Networks*, 22(5-6):544–557, July 2009.
[38] V. Vapnik, A. Vashist, and N. Pavlovitch. Learning using hidden information: Master-class learning. In F. F. Soulié, D. Perrotta, J. Piskorski, and R. Steinberger, editors, *NATO Science for Peace and Security Series, D: Information and Communication Security*, volume 19, pages 3–14. IOS Press, 2008.
[39] V. N. Vapnik. *The Nature of Statistical Learning Theory (Information Science and Statistics)*. Springer, 2nd edition, November 1999.
[40] J. Vesanto and E. Alhoniemi. Clustering of the self-organizing map. *Neural Networks, IEEE Transactions on*, 11(3):586–600, 2000.
[41] U. von Luxburg and **B. S.** David. Towards a statistical theory of clustering. In *PASCAL Workshop on Statistics and Optimization of Clustering*, 2005.
[42] F. Wilcoxon. Individual Comparisons by Ranking Methods. *Biometrics Bulletin*, 1(6):80–83, 1945.